\documentclass[conference,a4paper]{IEEEtran}
\usepackage{xcolor}
\usepackage{balance}

\usepackage{tabularx}

\usepackage[all=normal,paragraphs=tight,floats=normal,mathspacing=normal,wordspacing=tight,charwidths=tight,mathdisplays=normal,leading=tight]{savetrees}

\usepackage{amsmath,amssymb}

\usepackage{cite}
\ifCLASSINFOpdf
  \usepackage[pdftex]{graphicx}
\else
  \usepackage[dvips]{graphicx}
\fi
\usepackage{float}

\usepackage{amsmath}
\ifCLASSOPTIONcompsoc
 \usepackage[caption=false,font=normalsize,labelfont=sf,textfont=sf]{subfig}
\else
 \usepackage[caption=false,font=footnotesize]{subfig}
\fi
\usepackage{enumitem}
\setlist[itemize]{leftmargin=*}

\begin{document}
\title{ Efficient Frequency Selective Surface Analysis via End-to-End Model-Based Learning}

\author{\IEEEauthorblockN{
Cheima Hammami, 
Lucas Polo-López,   % 1st author, 1st affiliations
Luc Le Magoarou   % 2nd author, 2nd affiliations
   % 3rd author, 3rd affiliations
      % 4th author, 4th affiliations
}                                     % ...
%\\
\IEEEauthorblockA{% 1st affiliations
 Univ Rennes, INSA Rennes, CNRS, IETR-UMR 6164, Rennes, France}
\IEEEauthorblockA{
cheima.hammami@insat.ucar.tn; lucas.polo-lopez@insa-rennes.fr; luc.le-magoarou@insa-rennes.fr}
}



\maketitle

\begin{abstract}
This paper introduces an innovative end-to-end model-based deep learning approach for efficient electromagnetic analysis of high-dimensional frequency selective surfaces (FSS). Unlike traditional data-driven methods that require large datasets, this approach combines physical insights from equivalent circuit models with deep learning techniques to significantly reduce model complexity and enhance prediction accuracy. Compared to previously introduced model-based learning approaches, the proposed method is trained end-to-end from the physical structure of the FSS (geometric parameters) to its electromagnetic response (S-parameters). Additionally, an improvement in phase prediction accuracy through a modified loss function is presented. Comparisons with direct models, including deep neural networks (DNN) and radial basis function networks (RBFN), demonstrate the superiority of the model-based approach in terms of computational efficiency, model size, and generalization capability.
\end{abstract}


\section{Introduction}
Frequency selective surfaces (FSS) are a type periodic structures that respond differently to an incident electromagnetic wave depending on its frequency. More specifically, they can allow the propagation of certain frequency components while reflecting others. This behaviour is of great interest for many kinds of applications like EM-shielding, RCS reduction or polarization conversion among others \cite{Murugan2015, Hakim2024, Das2024, Liao2019, Molero2019}. To analyse and design FSSs, it is necessary to calculate their frequency response (i.e. which frequencies are reflected, and which are transferred). The most common way to express this frequency response is in terms of the scattering parameters (S-parameters), which relate incident and reflected power waves \cite[Ch.~4]{Pozar2011}. % This reference about S-parameters might be unnecessary if the paper is sent to a RF-related conference/journal.

Traditionally, FSSs have been analysed using equivalent circuits. This topic has been extensively covered by the literature and many sophisticated models do exist \cite{Costa2012,Medina2010,Mesa2015,Conde-Pumpido2022}. Although they are a great way to get an intuitive understanding of FSS behaviour while also being a very computationally efficient, equivalent circuits can quickly become extremely complex for relatively simple FSSs. Moreover, many of the circuit approaches impose requirements on the geometry of the FSS to ease the development of the circuit \cite{Mesa2018}, which limits the degrees of freedom that a designer can use to create highly performant FSS in the state of the art. Alternatively, purely numerical simulation based on techniques like the Finite-Difference Time-Domain Method \cite{Elsherbeni2015} or the Finite Element Method \cite{Jin2015} can characterize FSSs with any geometry. Nevertheless, these kinds of approaches tend to be computationally heavy, and therefore the simulation time can easily explode for relatively complex FSSs. This limits the application of numerical simulation when long optimization processes must be carried out.

Advancements in machine learning (ML) have shown promising results in the analysis of different types of periodic surfaces \cite{Goudos2022,Prado2018, Naseri2020, Lin2020,Fallah2023}. Nevertheless, classical data-driven methods, such as deep neural networks (DNNs) \cite{Koziel2022} or radial basis function networks (RBFNs), lack physical interpretability and often require large amounts of data \cite{Terayama2021}. This is of extreme importance since the training data is normally created via a numeric simulation, and therefore, generating a big train data set requires a great amount of computational resources.

Recently, works on model-based deep learning (DL) \cite{Shlezinger2023} have proposed to combine ML and domain-knowledge in order to achieve at the same time flexibility, data efficiency and interpretability. Model-based DL has been applied with great success to various tasks in wireless communications such as channel estimation \cite{Yassine2022,Chatelier2023}, precoding \cite{Lavi2023}, detection \cite{Samuel2017} or localization \cite{Mateos2023}. The domain of FSS analysis seems a very promising field for the application of model-based DL since the vast knowledge base on equivalent circuits can be used to circumvent the difficulties on dataset generation that classical ML approaches experience in this area.

{\bf \noindent Contributions.} To address the limitations of classical deep neural networks for the FSS analysis task, this paper presents a model-based deep learning approach that integrates the knowledge of equivalent circuit models \cite{Mesa2018} into the structure of the proposed neural network. It is a follow-up of a recent paper of ours \cite{Polo2024}, and extends it significantly in the following aspects:
\begin{itemize}
    \item \emph{End-to-end training.} The proposed method is fully trained from the geometric parameters to the S-parameters, which greatly enhances the accuracy of predictions.
    \item \emph{Phase prediction.} A new cost function is proposed that greatly improves the accuracy of phase predictions which is crucial in certain applications.
    \item \emph{Comparison with data-driven approaches.} The proposed method is rigorously compared with classical neural networks such as multilayer perceptrons (MLP) or radial basis function networks (RBFN), in terms of number of parameters, training time, smoothness of the predicted response and generalization capability, systematically showing the advantage of the proposed model-based approach over purely data-driven ones.
\end{itemize}

{\bf \noindent Dataset.} The dataset for training the model was generated using full-wave (FW) electromagnetic simulations in CST Microwave Studio. A parametric sweep over various design parameters resulted in 729 simulated S-parameter samples for different FSS geometries, with 9 examples taken between slot lengths (14.75--14.9 mm) and separations (8.79--10.3 mm). The S-parameters (S11, S21) were simulated over a frequency range of 6--16 GHz.

In the model-based approach, the dataset also includes initial equivalent circuit parameters derived from FW simulations \cite{garcia2012simplified}, providing an accurate starting point and reducing computational complexity.

%All the following improvements and comparisons were made using the same dataset as the existing model-based approach to train our enhanced model. The development framework employed was PyTorch, and the Adam optimization algorithm was used to train the models. 

{\bf \noindent Related work.} Advances in machine learning and electromagnetic simulation have opened new paths for efficient modeling of Frequency Selective Surfaces (FSS). Below, we review both traditional physical methods and modern machine learning approaches used in FSS analysis.

\begin{itemize}
    \item \emph{Physical Methods}: Historically, electromagnetic analysis of FSS has relied on well-established numerical methods, such as the Method of Moments (MoM) \cite{sultan2012method}, Finite Element Method (FEM) \cite{jin2008finite}, and Finite Difference Time Domain (FDTD) \cite{wnuk2005analysis} method. Additionally, the equivalent circuit method offers a simplified representation of FSS but remains limited in capturing complex interactions \cite{Costa2014,garcia2012simplified,mesa2018efficient}. While these methods provide highly accurate results, they are computationally intensive, especially for high-dimensional FSS structures and wide frequency ranges. The increasing demand for faster design iterations and large-scale optimization makes it impractical to rely solely on these traditional methods.
    
    \item \emph{Machine learning approaches}: In the last decade, machine learning techniques have been explored as an alternative for FSS modeling. These methods can be broadly categorized into two approaches:
    \begin{itemize}
        \item \emph{Direct (purely data-driven) methods}: These methods aim to learn a direct mapping between the geometric properties of the FSS and the electromagnetic response (S-parameters) \cite{goudos2022design,xiao2018multiparameter,jacobs2013two,koziel2021accurate}. While such models can predict the S-parameters without relying on physical knowledge, they often require large datasets and struggle to generalize outside the training distribution.
        
        \item \emph{Model-based methods}: These approaches incorporate physical knowledge into the machine learning pipeline \cite{Polo2024}. Instead of directly predicting the S-parameters, they first estimate intermediate physical quantities, such as equivalent circuit parameters, which are then used to compute the S-parameters. This hybrid approach reduces model complexity and leverages both data and physics to improve accuracy and generalization.
    \end{itemize}
\end{itemize}

%Our work is grounded in the latest approach, offering a new proposal for an improved model-based method that incorporates both machine learning and physical models, with an emphasis on efficient and accurate S-parameter prediction.

\begin{figure}[t]
\includegraphics[width=\columnwidth]{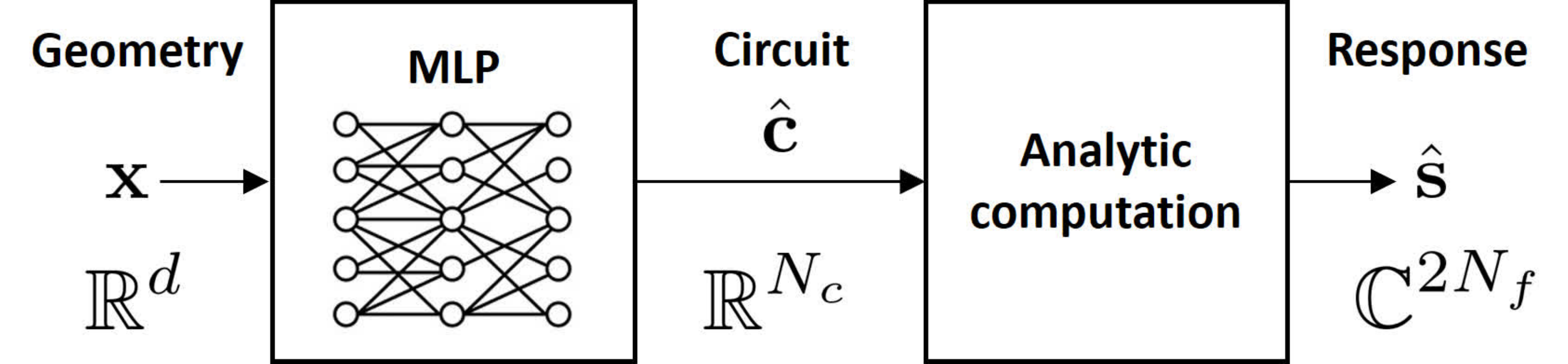}
\caption{Block diagram of the proposed model-based approach. The response of an FSS is predicted from its geometry via a circuital equivalent whose values are obtained through an MLP.}
\label{fig:schema}
\end{figure}

\section{Problem Formulation}

The goal of this work is to predict the S-parameters of an FSS based on its geometric and material properties. 
The input to the method consists of $d$ geometric features of the FSS gathered in the vector $$\mathbf{x} = [x_1,\dots, x_d] \in \mathbb{R}^d,$$ 
such as periodicity, patch size and substrate thickness. %and the frequency points $\mathbf{f}$ where the S-parameters need to be predicted.

%Thus, the input to the model is $\mathbf{X}$, representing the full set of features characterizing the FSS.

The output of the method is an estimated response, taking the form of S-parameters at a given set of $N_f$ frequencies. It is expected to be close to the actual response
    \[
    \mathbf{s} = [S_{11}(f_1), S_{21}(f_1), \dots, S_{11}(f_{N_f}), S_{21}(f_{N_f})] \in \mathbb{C}^{2N_f},
    \]
    where $S_{11}(f)$ and $S_{21}(f)$ are the reflection and transmission coefficients at frequency $f$.

    In the proposed model-based learning approach, S-parameters are not predicted directly from the geometric features. Indeed, $N_c$ \emph{equivalent circuit parameters}, denoted 
    $$\mathbf{c} = [c_1, \dots, c_{N_{c}}] \in \mathbb{R}^{N_c}$$
    are first estimated via a multilayer perceptron (MLP).
    Using these circuit parameters, the estimated S-parameters are then computed analytically via a physical model:
    \[
    \hat{\mathbf{s}} = f_{\text{phys}}(\hat{\mathbf{c}}),
    \]
    where $f_{\text{phys}}$ represents the physical equations derived from the equivalent circuit model.
    The whole workflow of the proposed method is illustrated in Figure \ref{fig:schema}.

{\noindent \bf Training.} In order to train the MLP, a dataset consisting of $N_s$ training samples (described in the introduction) is available, which takes the form
$$
\left\{\mathbf{x}^{(i)},\mathbf{c}^{(i)},\mathbf{s}^{(i)} \right\}_{i=1}^{N_s}.
$$
The ultimate objective is to minimize the difference between the predicted S-parameters $\mathbf{\hat{s}}$ and the ground truth $\mathbf{s}$ on average over the dataset obtained from FW simulations. To this end, the loss function used in \cite{Polo2024} is
\begin{equation}
\mathcal{L} = \frac{1}{N_f  N_s} \sum_{i=1}^{N_{s}} \sum_{j=1}^{N_{f}} \left| S_{21}^{(i)}(f_j) - \hat{S}_{21}^{(i)}(f_j) \right|.
\end{equation}
In this paper, a novel training strategy leading to much better results is proposed. It is presented in the next two sections.

%where $N_{\text{samples}}$ is the number of FSS samples, $N_{\text{freq}}$ is the number of frequency points, \( S_{21}(i, f_j) \) is the true transmission coefficient for the $i$-th sample at frequency \( f_j \), and \( \hat{S}_{21}(i, f_j) \) is the predicted transmission coefficient.

\begin{figure}[t]
\centering
\includegraphics[width=\columnwidth]{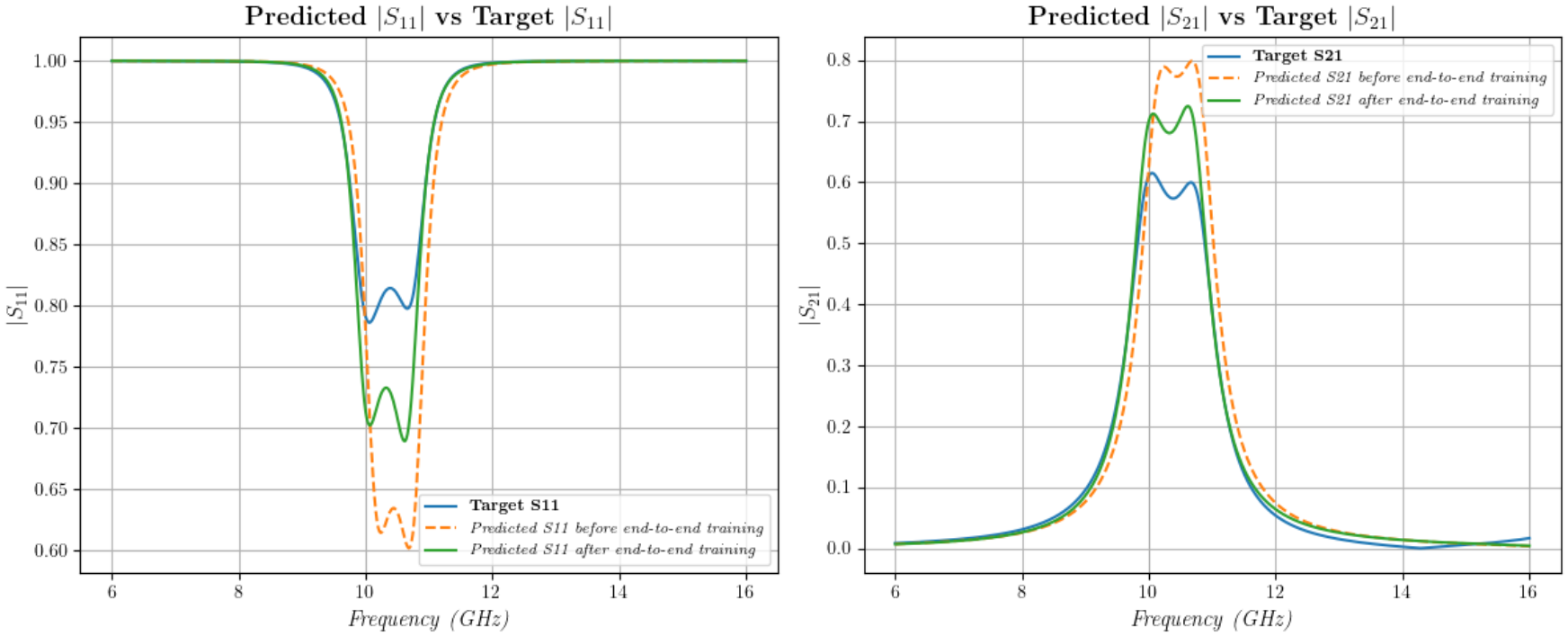}
\caption{Comparison of predictions before and after end-to-end training for a specific example.}
\label{fig:prediction_results}
\end{figure}

\begin{figure}[t]
\centering
\includegraphics[width=\columnwidth]{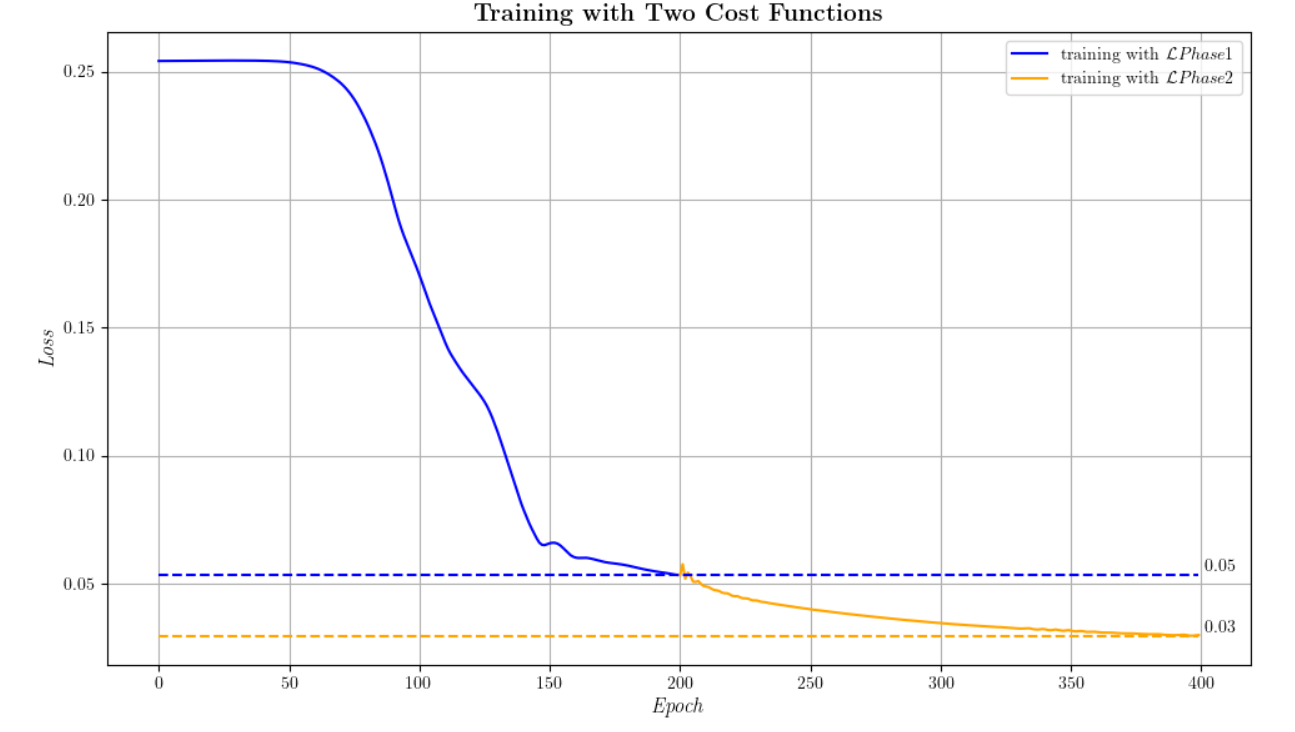}
\caption{Training curve showing the reduction in test error across the two training phases.}
\label{fig:training_curve}
\end{figure}

\section{End-to-end learning approach}

In the existing model-based framework, a multilayer perceptron (MLP) is trained to predict the equivalent circuit parameters of a frequency selective surface (FSS) based on its geometric configuration \cite{Polo2024}. After predicting the circuit parameters, the transmission matrix method \cite{Costa2014} is employed to compute the ABCD matrix of each FSS layer and its dielectric spacers. The ABCD matrices are then cascaded to form the total ABCD matrix, from which the S-parameters are derived.

However, in the original implementation, the cost function is focused solely on minimizing the error between the predicted and true circuit parameters, and not on the S-parameters, which are the final target. To improve the performance of the model, we propose a two-phase end-to-end training strategy. 

\subsection{Phase 1: Training on Circuit Parameters}

In the first phase, the MLP is trained to predict the equivalent circuit parameters. The cost function used in this phase is designed to minimize the mean absolute error (MAE) between the predicted circuit parameters $\hat{C_i}$ and the true circuit parameters $C_i$ for each sample. This phase allows the model to capture the underlying physics of the FSS through its equivalent circuit representation. The cost function for this phase is given by:

\begin{equation}
\mathcal{L}_{1} = \frac{1}{N_s} \sum_{i=1}^{N_s} \left\Vert \mathbf{c}^{(i)} - \hat{\mathbf{c}}^{(i)}\right\Vert_2^2
\end{equation}

where $N_s$ is the number of training samples, $c_i$ represents the true circuit parameters for the $i$-th sample, and $\hat{c_i}$ represents the predicted circuit parameters for the same sample.

This phase ensures that the MLP effectively learns to predict the circuit parameters based on the geometric configuration of the FSS, which simplifies the prediction process and reduces the overall complexity of the model.

\subsection{Phase 2: Training on S-Parameters}

After the initial training on the circuit parameters, the MLP is retrained in the second phase with a focus on improving the accuracy of the S-parameters. In this phase, we modify the cost function to minimize the  MAE between the predicted and true S-parameters, which are derived from the circuit parameters using the transmission matrix method. The goal here is to optimize the model's predictions directly for the S-parameters.

The updated cost function in Phase 2 is:

\begin{equation}
\mathcal{L}_{2} = \frac{1}{N_{\text{f}} \cdot N_{\text{s}}} \sum_{i=1}^{N_{\text{s}}} \sum_{j=1}^{N_{\text{f}}} \left| S_{21}(i, f_j) - \hat{S}_{21}(i, f_j) \right|
\end{equation}

where $N_{\text{s}}$ is the number of FSS samples, $N_{\text{f}}$ is the number of frequency points, \( S_{21}(i, f_j) \) is the true transmission coefficient for the $i$-th sample at frequency \( f_j \), and \( \hat{S}_{21}(i, f_j) \) is the predicted transmission coefficient.

\subsection{Results}

This two-phase strategy helps the model achieve higher prediction accuracy by first learning the physical structure through the circuit parameters and then refining its predictions based on the final S-parameters. It provides a more robust optimization process, leading to a significant reduction in the overall prediction error.

As shown in Figure \ref{fig:training_curve} of the training curve, we observe that the test error at the end of the second training phase has been reduced by half compared to the first phase. This resulted in more accurate predictions of the examples, as illustrated in Figure \ref{fig:prediction_results}, which compares the predictions before and after the end-to-end training.

\section{Phase-aware cost function}

To further improve the accuracy of phase predictions, we modified the cost function used during the second phase of training. The new cost function incorporates both S11 and S21 parameters to provide a more comprehensive evaluation of the model’s performance, especially in terms of phase prediction.

We observed that while the end-to-end training significantly reduced the error for S21, it did not yield the same improvement for S11. Despite the predicted magnitudes of S11 being as accurate as those of S21, as illustrated in Figure \ref{fig:prediction_results}, the prediction error for S11 remained relatively high. This can be explained by the relationship:

\begin{equation}
|S_{11}|^2 + |S_{21}|^2 = 1,
\end{equation}

which holds true for the magnitudes of these parameters, but no equivalent relationship exists that includes the phases. Therefore, it became necessary to adjust the cost function to also include S11.

The updated loss function is defined as:

\begin{align}
\text{Loss} = \frac{1}{N_{\text{f}} \cdot N_{\text{s}}} 
&\sum_{i=1}^{N_{\text{s}}} \sum_{j=1}^{N_{\text{f}}} 
\left( \hat{S}_{21}(i, f_j) - S_{21}(i, f_j) \right)^2 \nonumber \\
&+ \left( \hat{S}_{11}(i, f_j) - S_{11}(i, f_j) \right)^2
\end{align}

where \(\hat{S}_{11}\) and \(\hat{S}_{21}\) are the predicted S-parameters, and \(N_{\text{s}}\) and \(N_{\text{f}}\) denote the number of samples and frequency points, respectively.

\begin{figure}[t]
\centering
\includegraphics[width=\columnwidth]{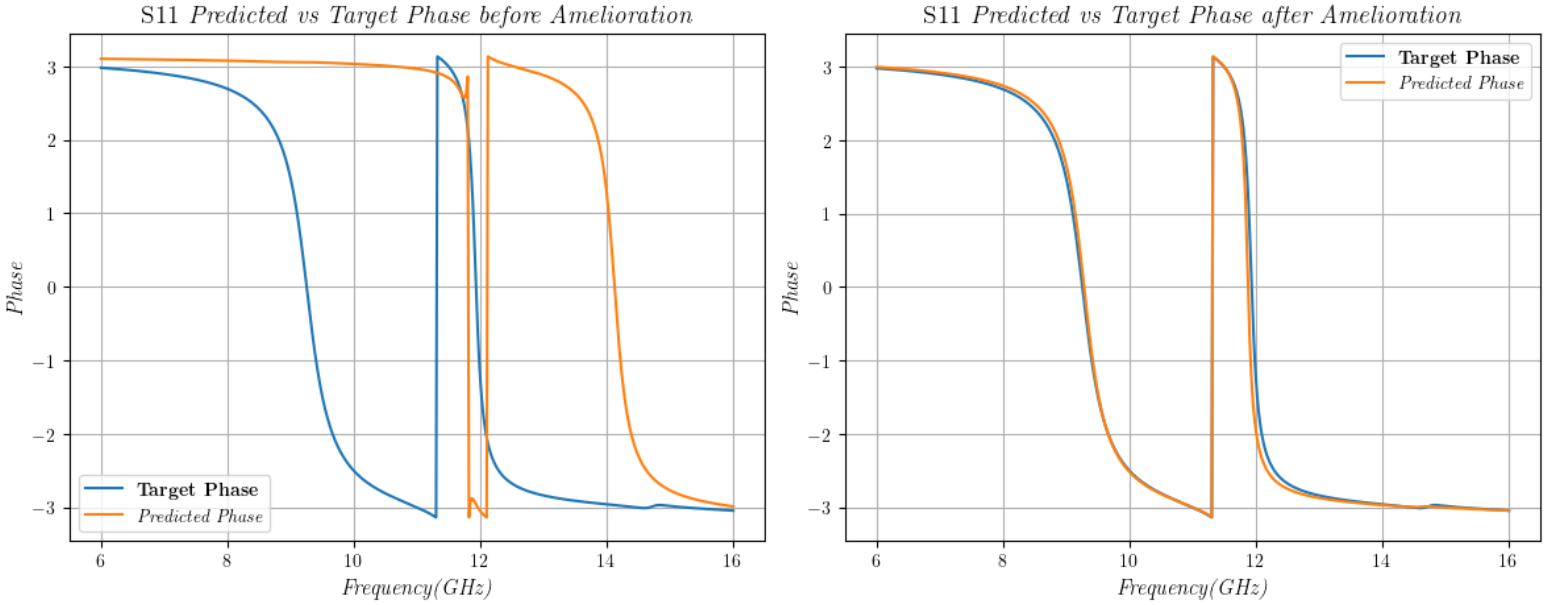}
\caption{Comparison of phase predictions before and after the adjustment of the cost function.}
\label{fig:phase_prediction}
\end{figure}

This adjustment significantly reduced the prediction error for S11, from 0.2 to 0.06, leading to a noticeable improvement in phase prediction accuracy, as demonstrated in Figure \ref{fig:phase_prediction} the phase predictions for a specific example before and after the improvement.

\section{Comparison with purely data-driven approaches}
We compared the model-based approach with two direct models: a deep neural network (DNN) and a radial basis function network (RBFN). The DNN architecture included fully connected layers of sizes 4, 8, 16, 32, 64, 128, 256, and 512, with dropout applied for regularization, while the RBFN used 200 centers. Each model was evaluated using a dataset of FSS geometries and their corresponding S-parameters.

\subsection{Prediction Quality}
The model-based approach demonstrated superior quality compared to both the DNN and RBFN. The integration of physical knowledge through the equivalent circuit model allowed for a better-guided learning process, minimizing underfitting and reducing the MAE. In contrast, the predictions from the DNN and RBFN models were noisy due to the lack of dependency in predictions between adjacent neurons and consequently between adjacent frequencies. Although the predictions from RBFNs were less noisy due to their Gaussian-shaped outputs resembling S-parameter responses, the model-based approach produced very smooth predictions that closely fit the target curves. This is illustrated in the following figures.

\begin{figure*}[htbp]
\centering
\includegraphics[width=\textwidth]{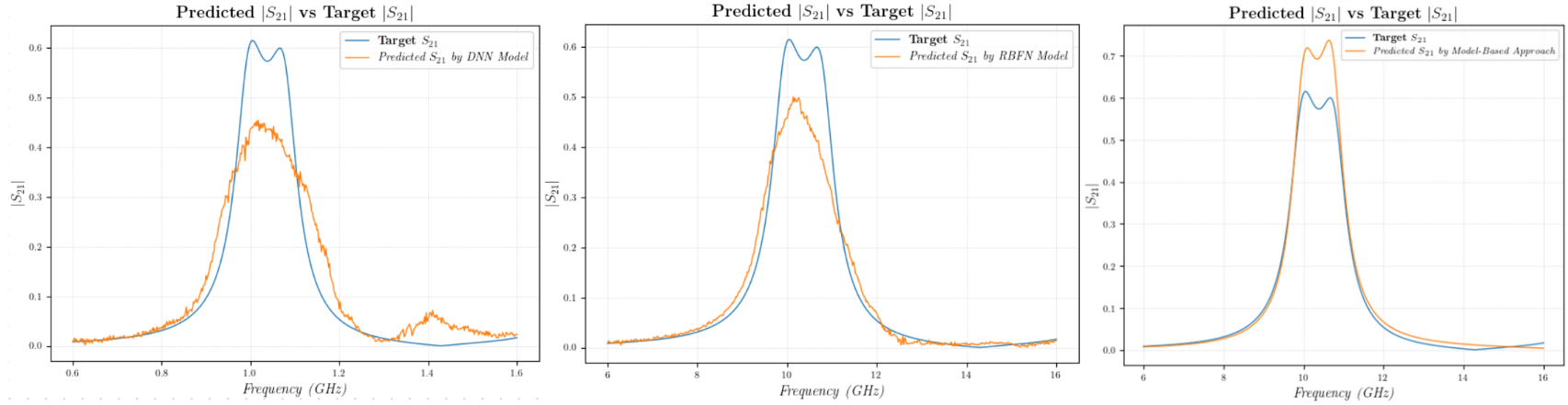}
\caption{Comparison of Prediction Quality Among Different Models}
\label{fig:prediction_qualities}
\end{figure*}

\begin{figure}[t]
\centering
\includegraphics[width=\columnwidth]{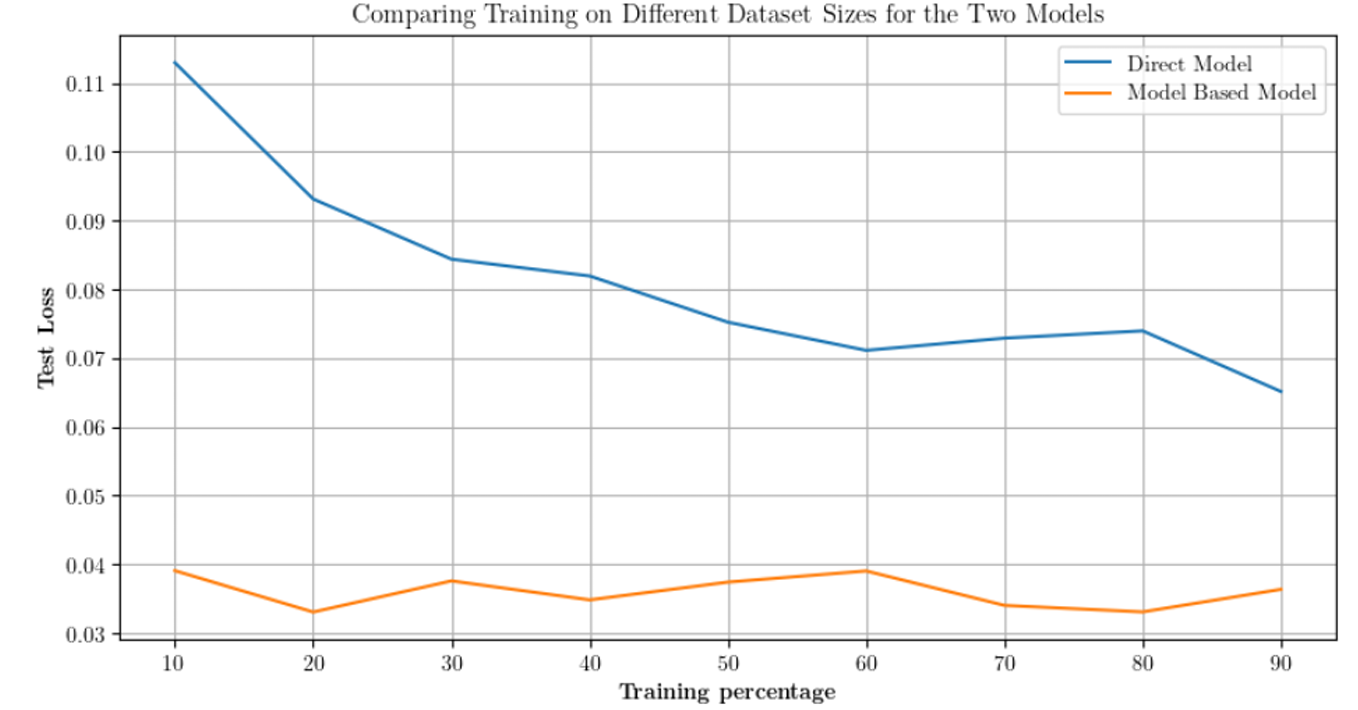}
\caption{Comparaison of the two approches' generalization capacities}
\label{fig:generalization_capcities}
\end{figure}

\subsection{Model Size and Training Time}

In Table \ref{tab:comparison_models}, we compare the performance of various models based on their prediction accuracy, training time, and the number of parameters. The model-based approach demonstrates superior accuracy compared to both the DNN and RBFN models. Its ability to leverage physical insights allows for a more effective learning process, which contributes to minimizing prediction errors. In contrast, while the DNN models showed improvement with the use of different activation functions, they still produced relatively higher prediction errors and required significantly longer training times. The RBFN, although slightly better in accuracy than some DNN variants, still could not match the efficiency and precision of the model-based method. Moreover, the model-based approach stands out with its remarkably lower parameter count, indicating a more streamlined architecture that effectively balances complexity and performance. This efficiency not only enhances prediction accuracy but also makes the model-based approach more practical for deployment in resource-constrained environments.

\begin{table}[H]
    \centering
    \caption{Comparison of Prediction Accuracy, Training Time, and Number of Parameters for Different Models}
    \resizebox{\columnwidth}{!}{%
    \begin{tabular}{|p{3cm}|p{1.5cm}|p{1.5cm}|p{1.5cm}|}
        \hline
        \textbf{Model} & \textbf{Test MAE} & \textbf{Training Time (s)} & \textbf{Number of Parameters} \\
        \hline
        DNN & 0.0689 & 1575 & 4,106,196 \\
        DNN with tanh activation & 0.0493 & 1621 & 1,225,044 \\
        RBFN & 0.0459 & 1439 & 805,204 \\
        Model-based & 0.0425 & 39.89 & 249 \\
        \hline
    \end{tabular}%
    }
    \label{tab:comparison_models}
\end{table}

\subsection{Generalization}
To assess the generalization capabilities of our model-based approach, we conducted experiments comparing it to direct models (DNN and RBFN). Both approaches were trained on a percentage of the dataset, then tested on the remaining unseen data, simulating real-world scenarios with new FSS geometries and frequency configurations.

The results, shown in Figure 6, highlight that the model-based approach consistently performed well across different dataset sizes, maintaining high accuracy even with limited data. In contrast, the direct models improved with larger datasets but remained less accurate overall.

This demonstrates a key advantage of the model-based approach: its ability to deliver reliable results even with small datasets. This is particularly valuable in practical applications where gathering large datasets is difficult. By leveraging prior knowledge and the inherent problem structure, the model-based method achieves superior generalization compared to traditional direct models.

\section{Conclusion}
This paper presented an end-to-end model-based deep learning approach for electromagnetic modeling of frequency selective surfaces. The proposed two-phase training strategy and the improvements in phase prediction accuracy demonstrate the effectiveness of this hybrid approach. Compared to direct models, the model-based method achieves better performance in terms of accuracy, efficiency, and generalization.
 Future work will explore further improvements by modeling more precisely the interactions between screens, which have so far been considered absent before refinement, using different representations through equivalent circuits, specifically \(\pi\) circuits \cite{mesa2018efficient}.

\bibliographystyle{IEEEtran}
\bibliography{biblio.bib}


\section*{Acknowledgment}
The work of L.~Le Magoarou is supported by the French national research agency (grant ANR-23-CE25-0013)

% \IEEEtriggeratref{7}
% \IEEEtriggercmd{\enlargethispage{-20cm}}


%\bibliographystyle{IEEEtran}

\end{document}